\pdfoutput=1
\documentclass{article}

\usepackage{arxiv}

\usepackage[utf8]{inputenc} 
\usepackage[T1]{fontenc}    
\usepackage{hyperref}       
\usepackage{url}            
\usepackage{booktabs}       
\usepackage{amsfonts}       
\usepackage{nicefrac}       
\usepackage{microtype}      
\usepackage{lipsum}
\usepackage{graphicx}
\graphicspath{ {./images/} }

\usepackage{amsmath}
\usepackage{amssymb}
\usepackage{mathtools}
\usepackage{amsthm}
\usepackage{tabularx}
\usepackage{rotating}
\usepackage{algorithm}
\usepackage{algorithmic}
\usepackage{diagbox}
\usepackage{makecell}

\title{Causal Discovery by Interventions via Integer Programming}

\author{
 Abdelmonem Elrefaey \\
  School of Coumputing and Augmented Intelligence\\
  Arizona State University\\
  Tempe, AZ 85281 \\
  \texttt{aelrefae@asu.edu} \\
   \And
 Rong Pan \\
  School of Coumputing and Augmented Intelligence\\
  University of Pittsburgh\\
  Arizona State University\\
  Tempe, AZ 85281 \\
  \texttt{rpan1@asu.edu} \\
}

\begin{document}
\maketitle
   \begin{abstract}
    Causal discovery is essential across various scientific fields to uncover causal structures within data. Traditional methods relying on observational data have limitations due to confounding variables. This paper presents an optimization-based approach using integer programming (IP) to design minimal intervention sets that ensure causal structure identifiability. Our method provides exact and modular solutions that can be adjusted to different experimental settings and constraints. We demonstrate its effectiveness through comparative analysis across different settings, demonstrating its applicability and robustness.
    \end{abstract}
    
    \section{Introduction}
    
    Causal discovery is a crucial endeavor in many scientific fields. Specifically, it focuses on revealing the causal structures within the data. Generally, causal discovery can be carried out through one of two data collection approaches -- observational data-based discovery and interventional or experimental data-based discovery. Most of past research employs observational methods, such as those using conditional independence tests, to provide valuable insights into causal structure. However, these methods have significant limitations, as they often face challenges from confounding variables and their inability to determine causality conclusively \cite{pearl2009causality,spirtes2000causation}. Observation-based causal discovery can most reveal a Directed Acyclic Graph (DAG) up to its Markov Equivalence Class (MEC) level, which means that there could be multiple different causal structures equivalent to fitting to the conditional independence revealed by the observational data alone \cite{verma1990equivalence}.
    
    The experimental approach, in contrast, provides a robust framework for discovering causal relationships by systematically manipulating variables and observing the effects \cite{fisher1935design}. The randomized controlled trial (RCT) is often considered the gold standard in this context, providing clear evidence of causality by isolating the effects of specific interventions \cite{rosenbaum2002observational}. However, traditional experimental designs presume a bipartite separation of variables into treatment and outcome categories (with possible covariates), which requires prior knowledge of which variables are potential causes (treatments) and which are potential effects (outcomes). They cannot adequately address the causal structure scenarios that do not fit this framework. For example, for a complex causal network, the conventional Design of Experiments (DOE) principles often do not provide useful guidance to the design of network interventions \cite{cox1958planning}.
    
    The primary goal of traditional DOE is to determine whether the treatment variable influences the outcome variable and, if so, to what extent. However, it does not offer advice on how to select treatment variables. Although bipartite separation is assumed in many studies given the condition of time-ordered information or limited intervention possibilities, there are numerous other situations where such restrictions are unwarranted (e.g., when multiple causal pathways exist between treatment and outcome variables).
    
    A more comprehensive framework for discovering causal structures -- causal Bayes nets -- has been developed in the fields of computer science and philosophy \cite{pearl2000causality,spirtes2000causation}. Unlike the bipartite structure assumed in DOE, causal Bayes nets accommodate general graphical networks (though this discussion is limited to acyclic networks). They represent causal relationships among variables by a DAG and a probability distribution on the variables in the DAG \cite{koller2009probabilistic}. This framework calls for new principles for intervention experimental designs and involves setting new design parameters such as selecting intervention and outcome variables, determining the type of intervention (structural or parametric), and the level of perturbation for each intervention variable. Moreover, the diversity of interventions should be planned to cover different aspects or levels of the variables of interest, including combinatorial interventions to explore joint effects and interactions while not exceeding cost limitations. Ensuring an adequate sample size is also a crucial design parameter for achieving sufficient statistical power, improving generalizability, and counting variations in outcome measures. The main focus in recent years has been on the number of intervention experiments required to reveal a causal structure under certain assumptions, as well as on deciding on which variables are to be intervened on in each experiment. The methodology by which many other design parameters remain open problems.

\section{Related Work}

In \cite{eberhardt_n_1_experiments} the authors prove the bounds for the minimal number of experiments required for causal discovery for one-intervened-variable or multiple-intervened-variable experiments.
Frederick Eberhardt's early work \cite{eberhardtPhD} has laid a solid groundwork for the field of interventional causal discovery. Two main types of interventions are discussed in his thesis: structural interventions and parametric interventions. Structural interventions break all incoming causal influences to an intervened variable, making it independent of its causes, thus it is effective in determining direct causal links from the intervened variable to other unintervened variables. Parametric interventions, on the other hand, modify the conditional probability distribution of the intervened variable without breaking incoming influences, preserving the original causal structure. This is necessary when structural intervention is impractical or unethical. His thesis explores the efficacy of fixed and adaptive \cite{Elahi_2024,Choo_2023,Choo_2023b,Sussex_2021} intervention strategies and discusses the efficiency of multiple simultaneous variable interventions.
Following Eberhardt's seminal work, \cite{eberhardt2012almost} specifies the minimal number of experiments needed to uniquely determine a causal graph from its observational MEC. The authors show that the number of required experiments is related to the largest clique in MEC. The OPTINTER algorithm proposed by the authors focuses on breaking down cliques quickly by selecting vertices in the largest cliques, but this algorithm is an NP-complete MAX-CUT problem.

\cite{hyttinen2013experiment} formalize the causal identifiability problem as variable pair conditions as follows:
\begin{itemize}
    \item Unordered Pair Condition (UPC): Ensures an experiment where one of any two variables is intervened upon and the other is passively observed.
    \item Ordered Pair Condition (OPC): Ensures an experiment where the first variable in any ordered pair is intervened upon and the second is passively observed.
    \item Covariance Condition (CC): Ensures an experiment where a pair of variables are passively observed. 
\end{itemize}
    The paper connects these conditions to combinatorial problems, using \emph{separating systems and cut-coverings} in graph theory following results in \cite{katona1966separating,renyi1961random}. This approach allows the construction of optimal experiment sets that satisfy the necessary conditions for causal discovery in the worst case (when the true DAG is fully connected).

Later work, such as that by \cite{hauser2014two} developed algorithms based on the optimal coloring of chordal graphs, requiring up to $ \log_2 (\chi) $ interventions to learn any causal graph, where $ \chi $ is the chromatic number of the chordal skeleton of the DAG.

\cite{shanmugam_small_interventions} focus on the problem of learning causal networks using interventions under Pearl’s Structural Equation Model with independent errors (SEM-IE). They aim to minimize the number of experiments needed to discover the causal directions of all edges in a causal graph. A key contribution is proving that any deterministic \emph{adaptive} algorithm needs to be a separating system to learn complete graphs in the worst case. They also present a novel separating system construction that is close to optimal and simpler than previous work in combinatorics. 

\cite{kocaoglu_cost_optimal} address the same problem with a cost consideration. They formulate the cost-optimal causal graph learning problem, aiming to design a set of interventions with a minimum total cost that can uniquely identify any causal graph given its skeleton. Their approach leverages the connection between graph-separating systems and proper colorings, although this connection is not widely known in the literature. The cost of an intervention is assumed to be a sum of costs assigned to each variable that is subject to manipulation which can be a special cost structure and may not be applicable in many real scenarios.

\cite{lindgren2018experimental} also explore the minimum cost intervention design problem, where the objective is to identify a set of interventions with a minimum total cost that can learn any causal graph given its essential graph representing its MEC. They demonstrate that this problem is NP-hard and provide a $ (2 + \epsilon) $-approximation algorithm using a greedy approach (a modification of \cite{kocaoglu_cost_optimal} proposed algorithm.
This algorithm is further extended to handle weighted cases, making it practical for real-world applications where intervention costs vary significantly.

\cite{ghassami_budgeted} tackles the budget-constrained causal discovery problem with non-adaptive experiments of size one. They formulate the problem as an optimization task to maximize the average number of resolved edge directions. 

    \section{Our Contribution}
    In this paper, we present a set of modular optimization models for interventional causal discovery, addressing the UPC, OPC, and CC outlined in \cite{hyttinen2013experiment}. We provide detailed mathematical formulations and demonstrate their effectiveness in identifying interventions that are minimally required to identify the causal structure involving a set of observable variables. This approach can be used to incorporate a wide variety of cost and number-of-variable-per-experiment constraints into the main formulation presented in this paper as will be demonstrated.
        
    \subsection{Optimization Approach}
    We formulate the interventional causal discovery problem into an Integer Programming (IP) formulation. The objective is to design a minimal set of interventions that satisfies the UPC, OPC, and CC, ensuring the entire causal structure among variables of interest is identifiable. 
    
    \subsubsection{Benefits of Using IP Optimization Models}
    Using IP (and optimization models in general) offers several benefits:
    \begin{itemize}
        \item \textbf{Exact Solutions}: IP models can provide exact solutions to intervention design problems, ensuring that the minimal set of interventions will be found.
        \item \textbf{Modularity}: The optimization models are modular, allowing for easy adjustments and extensions to incorporate additional constraints or prior knowledge such as the cost of each intervention. They can also accommodate various secondary objectives, such as minimizing the average number of variables manipulated per intervention or minimizing the maximum (worst case) number of variables in a single intervention.
        \item \textbf{Anytime Algorithm}: The Branch and Bound (B\&B) algorithm commonly used to solve IP problems can be viewed as an anytime algorithm; that is, it provides progressively better solutions the longer it runs. This allows for flexibility in computational time and resources, making it possible to obtain good solutions even under computational budget constraints. With a sufficient run time, it can provide \emph{all} solutions (intervention sets) that attain the minimum number of interventions necassary, which can be leveraged for various soft constraints in decision-making.
    \end{itemize}

\section{Problem Formulation}

Given a set of variables $\mathcal{V} = \{X_1, X_2, \ldots, X_N\}$, one of the goals of causal discovery is to determine the causal structure that describes how these variables influence each other. This structure is typically represented by a DAG, $G$, which is a tuple of a set of nodes and a set of directed edges connecting them. The nodes correspond to variables and directed edges represent direct causal relationships.

\subsection{Assumptions}
To ensure the identifiability of the causal structure, several assumptions need to be made about the underlying causal model:

\begin{itemize}
    \item \textbf{Acyclicity}: The graph $G$ is acyclic. This ensures that for any $X_i, X_j \in \mathcal{V}$, if $X_i$ causes $X_j$, then $X_j$ cannot reciprocally cause $X_i$. Formally, $G$ is a Directed Acyclic Graph (DAG).
    \item \textbf{Causal Sufficiency}: There are no unobserved common causes of two or more observed variables. Mathematically, for any pair of observed variables $X_i, X_j \in \mathcal{V}$, there does not exist an unobserved variable $U$ such that $U$ causally influences both $X_i$ and $X_j$.
    \item \textbf{Faithfulness}: The observed conditional independencies in the data correspond exactly to those implied by the causal graph. Formally, if $d$-separation in $G$ implies that $X_i \perp X_j \mid \mathbf{Z}$ (where $\perp$ denotes conditional independence), then $X_i \perp X_j \mid \mathbf{Z}$ must hold in the observed data as well.
        \item \textbf{Causal Markov Property}: Each variable is independent of its non-effects given its direct causes (parents) in the causal graph. Formally, for any $X_i \in \mathcal{V}$, $X_i \perp \text{NonDescendants}(X_i) \mid \text{PA}(X_i)$, where $\text{NonDescendants}(X_i)$ are all variables in $\mathcal{V}$ that are not descendants of $X_i$ in $G$.
    \item \textbf{Parametric Form}: The causal relationships follow a specific parametric form, often assumed to be linear. For instance, a linear structural equation model (SEM) can be expressed as:
    \[
    X_i = \sum_{X_j \in \text{PA}(X_i)} \beta_{ji} X_j + \epsilon_i,
    \]
    where $\text{PA}(X_i)$ denotes the set of parents of $X_i$, $\beta_{ji}$ are the linear coefficients, and $\epsilon_i$ are independent noise terms.
    \item \textbf{Independence Oracle}: An independence oracle is assumed to be available, which can correctly identify conditional independence relationships among the variables. Mathematically, the oracle can determine whether $X_i \perp X_j \mid \mathbf{Z}$ for any $X_i, X_j \in \mathcal{V}$ and any subset $\mathbf{Z} \subset \mathcal{V} \setminus \{X_i, X_j\}$.
\end{itemize}

\subsubsection{Achieving Structure Identifiability}

To achieve identifiability of the causal structure, certain combinations of UPC, OPC, and CC must be satisfied:
\begin{itemize}
    \item Assuming causal sufficiency, acyclicity, and faithfulness, the causal structure is identifiable if for any two variables $X_i, X_j \in \mathcal{V}$, two interventional experiments have been conducted where, either:
    \begin{itemize}
        \item the OPC holds for the pairs $(X_i, X_j)$ and $(X_j, X_i)$, or
        \item both the UPC and the CC hold for the pair $\{X_i, X_j\}$.
    \end{itemize}
    \item Without assuming causal sufficiency, acyclicity, or faithfulness, but assuming a linear data-generating model, the causal structure among observed variables is identifiable if the OPC holds for all ordered pairs of variables \cite{hyttinen2012causal,hyttinen2012learning, pmlr-v9-eberhardt10a}.
\end{itemize}

The focus of this work will be on the first case where no parametric form of the data model is assumed and when acyclicity, sufficiency, and faithfulness are met. Access to an independence oracle is also assumed.
    

In a multi-variable intervention experiment, any subset of variables in $\mathcal{V}$ can be manipulated while the remaining variables are observed. In such cases, \cite{frederick_eberhardt_number_2005} shows that at most half of all variables need to be intervened on to reveal all causal links. Interventions where more than half of the variables are intervened on are not optimal. Therefore, unless a limit on the number of variables that can be intervened on is imposed, only the interventions where up to $\lfloor|\mathcal{V}|/2\rfloor$ variables are manipulated are considered. 

\subsection{Notation}
We use the following notations in the optimization models:

\begin{itemize}
    \item $\mathcal{I}$: The set of all possible interventions.
    \item $\mathcal{K}$: The index set of all possible interventions.
    \item $\mathcal{S}_k$: The subset of variables subject to an intervention, $S_k \in \mathcal{I}$, where $k$ indicates the $k^{th}$ intervention.
    \item $I_k$: A binary decision variable indicating whether an intervention $\mathcal{S}_k$ is chosen ($I_k = 1$) or not ($I_k = 0$).
    \item $f_{ij}$: A binary decision variable indicating if there is a forward experiment constraint (see the identifiability section) for the pair $(X_i, X_j)$.
    \item $b_{ij}$: A binary decision variable indicating if there is a backward experiment constraint (see the identifiability section) for the pair $(X_i, X_j)$.
    \item $u_{ij}$: A binary decision variable indicating if there is a null experiment for the pair $(X_i, X_j)$.
\end{itemize}

\subsection{Objective Function}
The objective function for all base models is to minimize the number of chosen interventions:

\begin{equation}
\text{Minimize} \quad \sum_{k \in \mathcal{K}} I_k
\end{equation}

\subsection{Binary Constraint}
The binary constraint for all models is:

\begin{equation}
I_k \in \{0, 1\}, \quad \forall k \in \mathcal{K}
\end{equation}

\subsection{CC, UPC, and OPC Models}
We first present the formulation for satisfying the CC, UPC, and OPC individually.
\subsubsection{Covariance Condition}

The covariance condition ensures that for any pair of variables, there exists at least one intervention where neither of the variables is manipulated. This can be represented as:
\begin{align}
 \sum_{\substack{k \in \mathcal{K} \\ X_i \notin \mathcal{S}_k \land X_j \notin \mathcal{S}_k}} I_k \geq 1, \quad \forall \{X_i, X_j\} \subseteq \mathcal{V} \label{eq:cc_constraint}
\end{align}
Note that the null experiment, where no variables are manipulated (i.e. a passive observation of the system) is a trivial solution to this IP. 

\subsubsection{Unordered Pair Condition}

The unordered pair condition requires that for any pair of variables, there exists at least one intervention where one variable is affected, but the other is not. This is modeled as:

\begin{align}
 \sum_{\substack{k \in \mathcal{K} \\ (X_i \in \mathcal{S}_k \land X_j \notin \mathcal{S}_k) \\ \lor (X_i \notin \mathcal{S}_k \land X_j \in \mathcal{S}_k)}} I_k \geq 1, \quad \forall \{X_i, X_j\} \subseteq \mathcal{V} \label{eq:upc_constraint}
\end{align}

This ensures that for each pair $\{X_i, X_j\}$, there is an intervention where one is affected while the other remains unaffected.

\subsubsection{Ordered Pair Condition}

The ordered pair condition ensures that for each ordered pair of variables, there exist interventions where one variable is affected but not the other, and vice versa. This is formulated as two separate constraints:

\begin{align}
&\sum_{\substack{k \in \mathcal{K} \\ X_i \in \mathcal{S}_k \land X_j \notin \mathcal{S}_k}} I_k \geq 1, \quad \forall \{X_i, X_j\} \subseteq \mathcal{V} \label{eq:opc_constraint1} \\
& \sum_{\substack{k \in \mathcal{K} \\ X_i \notin \mathcal{S}_k \land X_j \in \mathcal{S}_k}} I_k \geq 1, \quad \forall \{X_i, X_j\} \subseteq \mathcal{V} \label{eq:opc_constraint2}
\end{align}

These constraints ensure that for any ordered pair $(X_i, X_j)$, there is an intervention where $X_i$ is affected and $X_j$ is not, and another intervention where $X_j$ is affected and $X_i$ is not.

\subsubsection{Identifiability}

Our DAG is a connected graph and any cyclic pattern is prohibited. Therefore, for a pair of variables ($X_i$, $X_j$), when both variables do not simultaneously appear in an intervention set, they should satisfy the following constraints to ensure their cause-effect direction is uniquely identifiable.

\begin{align}
& \sum_{\substack{k \in \mathcal{K} \\ X_i \in \mathcal{S}_k \land X_j \notin \mathcal{S}_k}} I_k \geq f_{ij}, \quad \forall \{X_i, X_j\} \subseteq \mathcal{V} \\
& \sum_{\substack{k \in \mathcal{K} \\ X_i \notin \mathcal{S}_k \land X_j \in \mathcal{S}_k}} I_k \geq b_{ij}, \quad \forall \{X_i, X_j\} \subseteq \mathcal{V} \\
& \sum_{\substack{k \in \mathcal{K} \\ X_i \notin \mathcal{S}_k \land X_j \notin \mathcal{S}_k}} I_k \geq u_{ij}, \quad \forall \{X_i, X_j\} \subseteq \mathcal{V} \\
& f_{ij} + b_{ij} + u_{ij} \geq 2, \quad \forall \{X_i, X_j\} \subseteq \mathcal{V} \\
& f_{ij}, b_{ij}, u_{ij} \in \{0,1\}, \quad \forall \{X_i, X_j\} \subseteq \mathcal{V} 
\label{eq:id_confirm_constraint}
\end{align}

In the inequalities above:
\begin{itemize}
    \item  Constraints (7) and (8) ensure that for each pair $\{X_i, X_j\}$, interventions are satisfying the forward and backward conditions respectively.
    \item Constraint (9) ensures there is at least one null experiment where both variables are unaffected.
    \item  Constraint (10) confirms that for each pair, at least two out of the three conditions (forward, backward, and null) are satisfied, ensuring identifiability.
\end{itemize}

These inequalities transform the UPC, OPC, and CC conditions into forward, backward, and null constraints. We define binary decision variables $f_{ij}$, $b_{ij}$, and $u_{ij}$ as follows:
\begin{itemize}
    \item $f_{ij}$ indicates whether there is a forward experiment constraint for the pair $(X_i, X_j)$. This corresponds to an intervention where $X_i$ is affected and $X_j$ is not.
    \item $b_{ij}$ indicates whether there is a backward experiment constraint for the pair $(X_i, X_j)$. This corresponds to an intervention where $X_j$ is affected and $X_i$ is not.
    \item $u_{ij}$ indicates whether there is a null experiment for the pair $(X_i, X_j)$. This corresponds to an intervention where neither $X_i$ nor $X_j$ is affected.
\end{itemize}

By requiring constraint (10), the IP model correctly integrates the identifiability condition. It ensures that either the OPC is satisfied, or the UPC and CC are satisfied (or both).

\subsubsection{The Set-Covering Problem}
The UPC, OPC, CC formulations in our interventional causal discovery framework can be viewed as instances of the Set Covering Problem (SCP), a well-known NP-hard problem in combinatorial optimization. The SCP aims to cover a universe of elements with the minimum number of subsets, which parallels our objective of identifying minimal intervention sets to ensure causal structure identifiability.
The SCP is NP-hard, meaning that finding an optimal solution is computationally infeasible for large instances \cite{Karp1972}. This complexity extends to our models, necessitating the exploration of efficient approximation techniques.

Given the generic formulation of intervention experimental design, our models can be extended to accommodate various practical considerations:
\begin{itemize}
    \item \textbf{Weighted Set Cover}: Incorporating intervention costs to minimize the total cost \cite{Chvatal1979}.
    \item \textbf{k-Set Cover}: Limiting the number of interventions to a fixed number $ k $ \cite{Hochbaum1996}.
    \item \textbf{Partial Set Cover}: Allowing for partial identifiability by covering a subset of causal relationships \cite{Slavik1996}.
\end{itemize}

For a large system, the following approximation algorithms can be employed to find near-optimal solutions:
\begin{itemize}
    \item \textbf{Greedy Algorithm}: Selects interventions iteratively, each time choosing the one that covers the most yet-uncovered pairs, achieving an $ O(\log n) $-approximation \cite{Johnson1974}.
    \item \textbf{Linear Programming (LP) Relaxation and Rounding}: If each element occurs in at most $ f $ sets, a solution can be found in polynomial time that approximates the optimum to within a factor of $ f $ using LP relaxation \cite{Vazirani2001}.

\end{itemize}

By framing the UPC, OPC, and CC models within the context of SCP, well-established optimization techniques can be leveraged to enhance the efficiency and effectiveness of causal discovery through interventions. Exact solutions to these optimization problems can guarantee optimality; conversely, approximation methods can provide flexibility and scalability, which is crucial for real-world applications.

\section{Extensions}

\subsection{Minimizing Cost of Interventions}

These IP models can be made to minimize the total cost of interventions instead of the number of interventions. This is achieved by changing the objective function to
\begin{equation}
\text{Minimize} \quad \sum_{k \in \mathcal{K}} C_kI_k
\end{equation}
where $C_k$ is the cost of intervention $I_k$. Since each $C_k$ can be set independently, the cost structure can be general and need not be limited to a linear function of the cost of intervening on individual variables. These modifications offer a substantial practical significance as minimizing the cost of interventions is a more realistic objective than the actual number of interventions. 

For example, consider the case of $\mathcal{V} = \{X_1, X_2, X_3, X_4\}$. Therefore, there are $11$ viable variable sets of up to size $\lfloor|\mathcal{V}|/2\rfloor = 2$ to be considered for an intervention, $\mathcal{I}$: 

\begin{itemize}
        \item $\mathcal{S}_0:\varnothing$
        \item $\mathcal{S}_1:\{X_1\}$
        \item $\mathcal{S}_2:\{X_2\}$
        \item $\mathcal{S}_3:\{X_3\}$
        \item $\mathcal{S}_4:\{X_4\}$
        \item $\mathcal{S}_5:\{X_1, X_2\}$
        \item $\mathcal{S}_6:\{X_1, X_3\}$
        \item $\mathcal{S}_7:\{X_1, X_4\}$
        \item $\mathcal{S}_8:\{X_2, X_3\}$
        \item $\mathcal{S}_9:\{X_2, X_4\}$
        \item $\mathcal{S}_{10}:\{X_3, X_4\}$
\end{itemize}

The theoretical lower bound for the necessary number of interventions required to identify a graph of $4$ variables is $\lfloor log_2(4)\rfloor + 1 = 3$ \cite{frederick_eberhardt_number_2005}. There are $\binom{11}{3} = 165$ possible 3-intervention solutions, but only $92$ of them can satisfy the identifiability conditions. Without cost considerations, the $92$ intervention experimental designs are equal solutions for this 4-variable causal discovery problem. However, if we consider the empty variable set (i.e. the null experiment, where all variables are passively observed) to have a zero cost and any other intervention experiment to have a one-unit cost, i.e.,  setting the IP objective function with ${C_0} = 0$ and ${C_k} = 1, \forall k \in \{1,\dots, 10\}$, yields only $12$ solutions that will attain the minimum cost. These solutions are:

\begin{enumerate}
        \item $(\mathcal{S}_0, \mathcal{S}_5, \mathcal{S}_6)$
        \item $(\mathcal{S}_0, \mathcal{S}_5, \mathcal{S}_7)$
        \item $(\mathcal{S}_0, \mathcal{S}_5, \mathcal{S}_8)$
        \item $(\mathcal{S}_0, \mathcal{S}_5, \mathcal{S}_9)$
        \item $(\mathcal{S}_0, \mathcal{S}_6, \mathcal{S}_7)$
        \item $(\mathcal{S}_0, \mathcal{S}_6, \mathcal{S}_8)$
        \item $(\mathcal{S}_0, \mathcal{S}_6, \mathcal{S}_{10})$
        \item $(\mathcal{S}_0, \mathcal{S}_7, \mathcal{S}_9)$
        \item $(\mathcal{S}_0, \mathcal{S}_7, \mathcal{S}_{10})$
        \item $(\mathcal{S}_0, \mathcal{S}_8, \mathcal{S}_9)$
        \item $(\mathcal{S}_0, \mathcal{S}_8, \mathcal{S}_{10})$
        \item $(\mathcal{S}_0, \mathcal{S}_9, \mathcal{S}_{10})$

\end{enumerate}

\begin{figure}[t]
    \centering
    \includegraphics[width=1.\columnwidth]{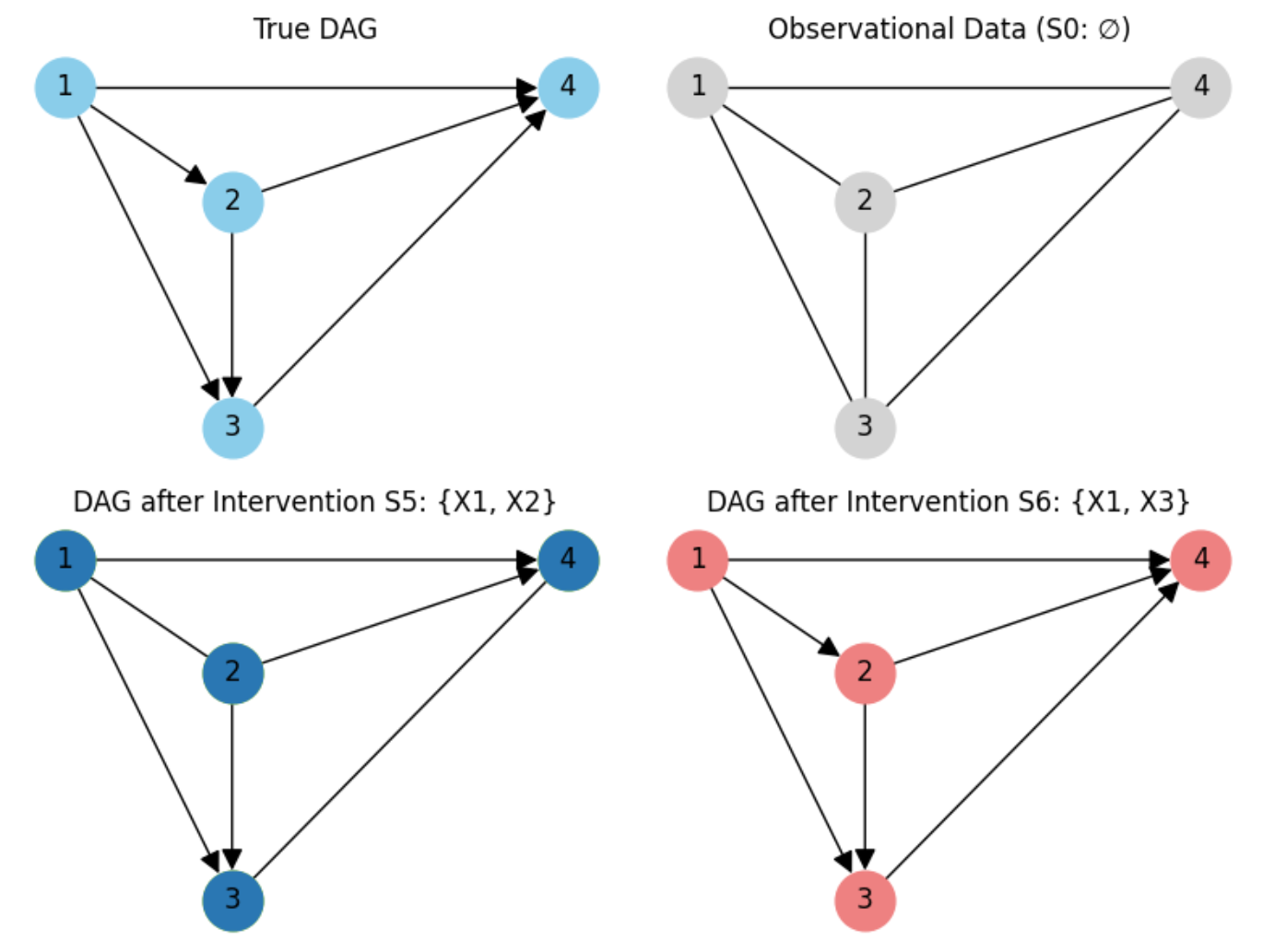} 
    \caption{True DAG inference through interventions from solution 1}
    \label{fig1}
    \end{figure}
    
Figure \ref{fig1} illustrates the evolution of inference of applying the intervention sets of solution 1. The True DAG outlines the actual causal relationships among four variables (1, 2, 3, and 4), where node 1 directly influences nodes 2, 3, and 4, node 2 influences nodes 3 and 4, and node 3 directly influences node 4. The Observational Data DAG, inferred from purely observational data without interventions, shows undirected edges between all nodes due to dependencies, but lacks clear causal directions. After intervention $\mathcal{S}_5$, where nodes 1 and 2 are intervened in simultaneously, the DAG reveals that both node 1 and node 2 directly affect nodes 3 and 4, but the direct relationship between nodes 1 and 2 cannot be determined as they are manipulated simultaneously. Likewise, the directionality between nodes 3 and 4 cannot be inferred. After intervention $\mathcal{S}_6$, where interventions are applied to nodes 1 and 3, the DAG is further clarified that node 1 directly affects node 2 and node 3 affects node 4. A similar inference can be made by choosing any of the other $11$ solutions.

\subsection{Limited number of variables that can be subject to an intervention simultaneously}

When there is a limit on the maximum number of variables that can be intervened simultaneously in an interventional experiment, $k_{max}$, the IP model can incorporate this requirement by defining $\mathcal{I}$ to only include the sets of
$|\mathcal{S}_k| \leq k_{max}$. The branch-and-bound algorithm will only seek to minimize the number of interventions necessary (or the total cost of interventions) over the reduced space of possible interventions which ensures that the limit is satisfied. Table \ref{table:id} shows the change in the number of interventions necessary to achieve identifier of the graph as a function of $N$ and $k_{max}$. As one can see, for example, where there are 9 variables in the system, if only one variable can be intervened on at a time then at least 8 interventional experiments are needed to identify the system; but this number can be reduced to 6 when two variables can be simultaneously intervened on, and it can be further reduced to 4 when three or four variables can be simultaneously intervened on. Similar results for the CC, UPC, and OPC can be found in the Appendix.

\begin{table}[h]
\centering
\resizebox{\columnwidth}{!}{
\begin{tabular}{|c|c|c|c|c|}
\hline
\diagbox[width=1.5cm, height=1.5cm]{\textbf{$N$}}{\textbf{$k_{max}$}} & 1 & 2 & 3 & 4 \\
\hline
2  &  $2\ (<0.1)$ &  - &  - &  - \\
\hline
3  &  $2\ (<0.1)$ &  - &  - &  - \\
\hline
4  &  $3\ (<0.1)$ &  $3\ (<0.1)$ &  - &  - \\
\hline
5  &  $4\ (<0.1)$ &  $3\ (<0.1)$ &  - &  - \\
\hline
8  &  $7\ (<0.1)$ &  $5\ (0.1)$ &  $4\ (0.4)$ &  $4\ (0.9)$ \\
\hline
9  &  $8\ (<0.1)$ &  $6\ (0.8)$ &  $4\ (0.2)$ &  $4\ (1.9)$ \\
\hline
16 &  $15\ (<0.1)$ &  $10\ (>600.0)$ &  $8\ (>600.0)$ &  $6\ (>600.0)$ \\
\hline
17 &  $16\ (<0.1)$ &  $11\ (>600.0)$ &  $8\ (>600.0)$ &  $7\ (>600.0)$ \\
\hline
\end{tabular}
}
\caption{Model solution to satisfying the Identifiability conditions for $N$ variables and a $k_{max}$ limit on the number of variables that can be manipulated simultaneously. The number in parenthesis is solution time in Seconds. Model minimizes objective function (1) subject to constraints: (2) and (7)-(11). The solution time was capped at 10 minutes. Scenarios, where the solution exceeded this time, may not be optimal.}
\label{table:id}
\end{table}

\subsection{Secondary objectives} \cite{hyttinen2013experiment} provide algorithms that not only yield a set of interventions that meet the theoretical bounds of minimum necessary interventions to meet the UPC or OPC respectively but also construct these interventions while minimizing either the average or the maximum intervention size. The IP approach can likewise be used to find the minimum number of interventions and additionally optimize another practically significant objective. As discussed before, there often exist many optimal solutions to the main objective function, choosing a solution that can also optimize a secondary objective is thereof desirable. \cite{eberhardtPhD} presents such a case when the number of variables,  $|\mathcal{V}|$, is 8 for which the theoretical lower bound of number of interventions required (assuming multi-variable interventions are permissible) is 4. Multiple configurations of intervention sizes can attain this bound, such as 4 experiments with intervention sizes of $(4,4,4,0)$, or of $(3,3,3,1)$.
The identifiability IP model can be solved to obtain all configurations for any $N$ and $k_{max}$. Examples are shown in Table \ref{table:config}.

\begin{table}[h]
\tiny
\centering
\resizebox{\columnwidth}{!}{
\begin{tabular}{|c|c|c|c|}
\hline
\diagbox[width=1.25cm, height=.5cm]{\textbf{$N$}}{\textbf{$k_{max}$}} & 1 & 2 & 3 \\
\hline
2  & \makecell{(0,1)\\ (1,1)} &  - &  - \\
\hline
3  &  \makecell{(1,1)} &  - &  - \\
\hline
4  &  \makecell{(1,1,1)} & \makecell{(0,2,2)\\ (1,1,1)\\ (1,1,2)\\ (1,2,2)\\  (2,2,2)} &  - \\
\hline
5  &  \makecell{(1,1,1,1)} & \makecell{(1,2,2)\\ (2,2,2)} &  - \\
\hline
6  &  \makecell{(1,1,1,1,1)} & \makecell{(1,1,2,2)\\ (1,2,2,2)\\ (2,2,2,2)} & \makecell{(2,2,3)\\ (2,3,3)\\ (3,3,3)} \\
\hline
\end{tabular}
}
\caption{Possible solution configurations to attain identifiability without secondary objectives.}
\label{table:config}
\end{table}

On the other hand, setting the minimization of the average intervention size as a secondary objective will yield the configurations in Table \ref{table:config+}. The size of an intervened variable set is often of practical importance in implementation, as a larger size can greatly increase the complexity of an experiment. Therefore, a smaller average size of interventional experiments is more desirable.

\begin{table}[h]
\tiny
\centering
\resizebox{\columnwidth}{!}{
\begin{tabular}{|c|c|c|c|}
\hline
\diagbox[width=1.25cm, height=0.5cm]{\textbf{$N$}}{\textbf{$k_{max}$}} & 1 & 2 & 3\\
\hline
2  & (0,1) &  - &  -\\
\hline
3  &  (1,1) &  - &  -\\
\hline
4  &  (1,1,1) &   (1,1,1) &  -\\
\hline
5  &   (1,1,1,1) &  (1,2,2) &  -\\
\hline
6  &   (1,1,1,1,1) &  (1,1,2,2) &  (2,2,3) \\
\hline
\end{tabular}
}
\caption{Possible solution configurations to attain identifiability with a secondary objective of minimizing average intervention size.}
\label{table:config+}
\end{table}

By utilizing our IP framework, it is easy to optimize secondary objectives such as the minimal size of intervened variable sets. Our final solutions can be refined to adhere to some soft decision-making requirements. 

\section{Conclusion}
In this paper, we introduce an optimization-based approach to causal discovery through interventions, leveraging integer programming to design the minimal number of intervention sets that ensure the identifiability of causal structures. Our method addresses the limitations of traditional observational methods, offering exact and modular solutions adaptable to various experimental settings and constraints.

We have demonstrated the effectiveness of our approach through detailed mathematical formulations and comparative analyses, highlighting its robustness and applicability across different scenarios. The flexibility of our IP models allows for easy integration of additional constraints, such as intervention costs, and supports the optimization of secondary objectives, including minimizing the average or maximum number of variables to be manipulated per interventional experiment. Our approach can be extended to various practical considerations, incorporating weighted set cover, k-set cover, and partial set cover formulations, to handle real-world problem complexities. Its ability to provide exact solutions and/or to progressively improve approximated solutions via the branch-and-bound algorithm enhances its utility under computational and budget constraints.

The proposed IP-based intervention design framework advances the field of causal discovery by offering a comprehensive and flexible solution for designing interventional experiments. Future research directions include enhancing computational efficiency, integrating prior structural knowledge, and broadening the framework’s applicability to more complex and realistic scenarios.

\bibliography{references.bib}
\bibliographystyle{unsrt}  

\newpage

    \appendix
    \section{Appendix}
    The appendix includes tabulated results on the minimum number of interventions necessary to satisfy the CC, UPC, and OPC respectively. 
    \begin{table}[ht]
    \centering
    \resizebox{\columnwidth}{!}{
    \begin{tabular}{|c|c|c|c|c|}
    \hline
    \diagbox[width=1.5cm, height=1cm]{\textbf{$N$}}{\textbf{$k_{max}$}} & \textbf{1} & \textbf{2} & \textbf{3} & \textbf{4} \\
    \hline
    2  &  $1\ (<0.1)$ &  $1\ (<0.1)$ &  $1\ (<0.1)$ &  $1\ (<0.1)$ \\
    \hline
    3  &  $1\ (<0.1)$ &  $1\ (<0.1)$ &  $1\ (<0.1)$ &  $1\ (<0.1)$ \\
    \hline
    4  &  $1\ (<0.1)$ &  $1\ (<0.1)$ &  $1\ (<0.1)$ &  $1\ (<0.1)$ \\
    \hline
    5  &  $1\ (<0.1)$ &  $1\ (<0.1)$ &  $1\ (<0.1)$ &  $1\ (<0.1)$ \\
    \hline
    8  &  $1\ (<0.1)$ &  $1\ (<0.1)$ &  $1\ (<0.1)$ &  $1\ (<0.1)$ \\
    \hline
    9  &  $1\ (<0.1)$ &  $1\ (<0.1)$ &  $1\ (<0.1)$ &  $1\ (<0.1)$ \\
    \hline
    16 &  $1\ (<0.1)$ &  $1\ (<0.1)$ &  $1\ (<0.1)$ &  $1\ (<0.1)$ \\
    \hline
    17 &  $1\ (<0.1)$ &  $1\ (<0.1)$ &  $1\ (0.1)$ &  $1\ (0.2)$ \\
    \hline
    \end{tabular}
    }
    \caption{Model solution to satisfying the CC for $N$ variables and a $k_{max}$ limit on the number of variables that can be manipulated simultaneously. The number in parenthesis is solution time in Seconds. The model minimizes objective function (1) subject to constraints: (2) and (3).}
    \label{table:ta1}
    \end{table}
    
    \begin{table}[ht]
    \centering
    \resizebox{\columnwidth}{!}{
    \begin{tabular}{|c|c|c|c|c|}
    \hline
    \diagbox[width=1.25cm, height=1cm]{\textbf{$N$}}{\textbf{$k_{max}$}} & \textbf{1} & \textbf{2} & \textbf{3} & \textbf{4} \\
    \hline
    2  &  $1\ (<0.1)$ &  $1\ (<0.1)$ &  $1\ (<0.1)$ &  $1\ (<0.1)$ \\
    \hline
    3  &  $2\ (<0.1)$ &  $2\ (<0.1)$ &  $2\ (<0.1)$ &  $2\ (<0.1)$ \\
    \hline
    4  &  $3\ (<0.1)$ &  $2\ (<0.1)$ &  $2\ (<0.1)$ &  $2\ (<0.1)$ \\
    \hline
    5  &  $4\ (<0.1)$ &  $3\ (<0.1)$ &  $3\ (<0.1)$ &  $3\ (<0.1)$ \\
    \hline
    8  &  $7\ (<0.1)$ &  $5\ (<0.1)$ &  $4\ (0.1)$ &  $3\ (<0.1)$ \\
    \hline
    9  &  $8\ (<0.1)$ &  $6\ (0.2)$ &  $4\ (0.5)$ &  $4\ (1.7)$ \\
    \hline
    16 &  $15\ (<0.1)$ &  $10\ (1.3)$ &  $8\ (8.1)$ &  $6\ (24.6)$ \\
    \hline
    17 &  $16\ (<0.1)$ &  $11\ (2.4)$ &  $8\ (8.4)$ &  $7\ (50.0)$ \\
    \hline
    \end{tabular}
    }
    \caption{Model solution to satisfying the UPC for $N$ variables and a $k_{max}$ limit on the number of variables that can be manipulated simultaneously. The number in parenthesis is solution time in Seconds. The model minimizes objective function (1) subject to constraints: (2) and (4).}
    \label{table:ta2}
    \end{table}

    \begin{table}[ht]
    \centering
    \resizebox{\columnwidth}{!}{
    \begin{tabular}{|c|c|c|c|c|}
    \hline
    \diagbox[width=1.5cm, height=1cm]{\textbf{$N$}}{\textbf{$k_{max}$}} & \textbf{1} & \textbf{2} & \textbf{3} & \textbf{4} \\
    \hline
    2  &  $2\ (<0.1)$ &  $2\ (<0.1)$ &  $2\ (<0.1)$ &  $2\ (<0.1)$ \\
    \hline
    3  &  $3\ (<0.1)$ &  $3\ (<0.1)$ &  $3\ (<0.1)$ &  $3\ (<0.1)$ \\
    \hline
    4  &  $4\ (<0.1)$ &  $4\ (<0.1)$ &  $4\ (<0.1)$ &  $4\ (<0.1)$ \\
    \hline
    5  &  $5\ (<0.1)$ &  $5\ (<0.1)$ &  $5\ (<0.1)$ &  $5\ (<0.1)$ \\
    \hline
    8  &  $8\ (<0.1)$ &  $8\ (<0.1)$ &  $6\ (0.1)$ &  $5\ (0.1)$ \\
    \hline
    9  &  $9\ (<0.1)$ &  $9\ (<0.1)$ &  $6\ (0.1)$ &  $5\ (0.1)$ \\
    \hline
    16 &  $16\ (<0.1)$ &  $16\ (0.2)$ &  $11\ (1.6)$ &  $8\ (5.1)$ \\
    \hline
    17 &  $17\ (<0.1)$ &  $17\ (0.3)$ &  $12\ (2.3)$ &  $9\ (14.3)$ \\
    \hline
    \end{tabular}
    }
    \caption{Model solution to satisfying the OPC for $N$ variables and a $k_{max}$ limit on the number of variables that can be manipulated simultaneously. The number in parenthesis is solution time in Seconds. Model minimizes objective function (1) subject to constraints: (2) and (5)-(6).}
    \label{table:ta3}
    \end{table}



\end{document}